\documentclass[10pt,twocolumn,letterpaper]{article}

\usepackage{cvpr}
\usepackage{times}
\usepackage{epsfig}
\usepackage{graphicx}
\usepackage{amsmath}
\usepackage{amssymb}

\cvprfinalcopy 

\def\cvprPaperID{****} 

\ifcvprfinal\fi
\begin{document}

\title{ Merging Satellite Measurements of Rainfall\\Using Multi-scale Imagery Technique}

\author{Seyed Hamed Alemohammad, Dara Entekhabi\\
Department of Civil and Environmental Engineering, MIT\\
{\tt\small hamed\_al@mit.edu, darae@mit.edu}
\and}

\maketitle
\thispagestyle{empty}

\begin{abstract}
Several passive microwave satellites orbit the Earth and measure rainfall. These measurements have the advantage of almost full global coverage when compared to surface rain gauges. 
However, these satellites have low temporal revisit and  missing data over some regions.
Image fusion is a useful technique to fill in the gaps of one image (one satellite measurement) using another one. 
The proposed algorithm uses an iterative fusion scheme to integrate information from two satellite measurements. 
The algorithm is implemented on two datasets for 7 years of half-hourly data. The results show significant improvements in rain detection and rain intensity in the merged measurements.

\end{abstract}

\section{Introduction}

Satellite monitoring of rainfall has the advantage of almost global areal coverage compared
with ground measurements (gauge and radar). Satellite rainfall retrievals have enormous
potential benefits as input to hydrological and agricultural models because of their real time
availability, low cost, and full spatial coverage~\cite{Grimes08}. One category of these measurements 
are carried out by Passive Microwave (PMW) sensors in Low Earth Orbit (LEO) platforms.
PMW measurements of rainfall have their own unique advantages as well as shortcomings. 
These sensors have a low temporal revisit (usually two times a day) because of their orbital characteristics. Moreover, 
in some cases they have missing data over some regions where retrieval is confounded by other influences. 
Therefore, in order to obtain reasonably accurate rainfall estimation and with a good spatial 
and temporal sampling resolution, useful for studies of climate change and large-scale hydrological processes,
 it is necessary to merge different types of measurements.

It is not possible to fill in the gaps of one measurement (hereafter, called image) using another one because none of these images are perfect measurements, and this will cause discontinuities in the final image. In addition, the comparison of results with the results of an interpolation algorithm (which produces a merged image by interpolating between two input images) shows that a simple interpolation is not enough to merge the two images, and the results are not accurate.

In order to produce a better estimation of the rainfall, researchers have tried to combine
different types of rainfall measurement. Several models have been developed that combines satellite measurements
with ground-base measurements, and provide estimation of rainfall for both missing pixels and for the times that 
there has not been a satellite measurement. Some of these models include: Global Precipitation Climatology Project 
(GPCP)~\cite{Lensky08}, Precipitation Estimation from Remotely Sensed Information using Artificial Neural Networks 
(PERSIANN)~\cite{Hsu08}, Climate prediction center MORPHing method (CMORPH)~\cite{Joyce10}, and Climate prediction 
center Merged Analysis of Precipitation (CMAP)~\cite{Xie07}. Shen \etal (2010) investigated the errors and biases of these merged measurements~\cite{Shen10} and showed that they are not reliable in some regions. Here, I will only focus on improving the retrievals when there are at least two satellite retrievals available over one region. However, this method can be applied to any two sets of rainfall measurement to fill in the missing regions.

One of the methods that can be used for image fusing is multi-resolution pyramids method~\cite{Adelson84}. This method is based on fusing the images in a transformed domain, created by decomposing images into sub-images. In this paper, I propose an algorithm for merging two rainfall images using the multi-resolution pyramid method and an interpolation scheme. The proposed algorithm has two section: \emph{Texture Production} and \emph{Shape Production}. In order to produce the texture, it uses the multi-resolution pyramid method, and for producing the shape, it uses an interpolation scheme. The results show that the distinction between shape and texture in the algorithm improves the results both in detecting the rainfall and in producing the accurate rain intensity.

The paper has four section. First, the data sets that were used are described. Then, the algorithm is presented in detail. Next, results are provided with comparisons with other methods. Last, a discussion on the results is presented.
\begin{figure}
	\begin{center}
		  \includegraphics[width=0.8\linewidth]{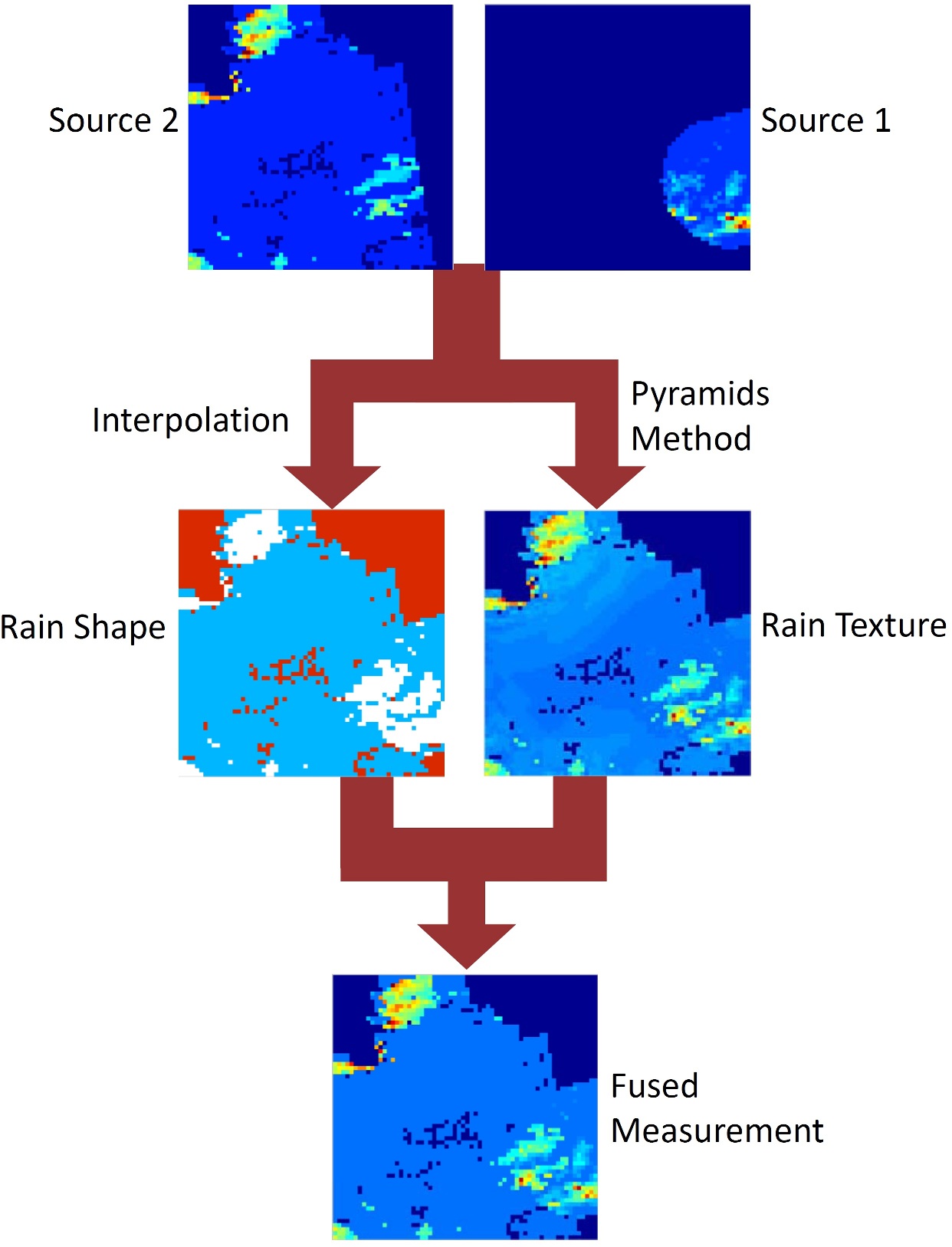}
		   \caption{Procedure of the proposed algorithm. Rain Shape is an image of [-1, 0, 1]. Other images show rain intensity (from blue to red the intensity increases from zero to its maximum; dark blue are the missing regions)}
		\label{approach}
	\end{center}
\end{figure}

\subsection{Data Sets}
\label{dataset}
The satellite measurements of rainfall for this study are obtained from National Oceanic and Atmospheric Administration (NOAA) 
satellites 15 and 16 (NOAA-15 and NOAA-16) and Tropical Rainfall Measuring Mission (TRMM). 
The former is produced by Microwave Surface and Precipitation Products System (MSPPS) in NOAA and 
the latter is produced by Goddard Earth Sciences Data and Information Services Center in 
National Aeronautics and Space Administration (NASA). The rainfall rates are from the AMSU-B 
sensors on board the NOAA-15 and NOAA-16 and the TRMM Microwave Imager (TMI) on board the TRMM satellite. 
An independent set of surface-based radar measurements produced by NEXt-Generation RADar Network 
(NEXRAD-IV) is assumed as the true measurement and used to evaluate the results of the model. 
This assumption can be further investigated by incorporating the errors of the radar measurements.

For this study, all the data are mapped to a common spatial gird of $0.25^{\circ} latitude \times 0.25^{\circ} longitude$ covering the Central U.S. region ranging $26.10^{\circ}-42.10 ^{\circ} N$ and $107.85^{\circ}-91.85^{\circ} W$. The period of study was from Jan. 2004 until Dec. 2010. All the data are half-hourly measurements.

In order to obtain a good dataset, I selected images that had a reasonable number (at least 40) of non-missing pixels. Imposing this criteria, a set of 219 coincident images for each of the NOAA-16 and TRMM, and a set of 221 coincident images for each of the NOAA-15 and TRMM was created. It should be noted that each sample is a $64 \times 64$ pixel image.

\section{Methodology}

The procedure for producing the merged images consists of two steps: \emph{texture} simulation and \emph{shape} simulation. Here, texture is the rain rate (or intensity) and shape is the rain support (\ie rain / no rain areas.) The complete procedure is illustrated in Figure~\ref{approach}. Using the two input images, a texture and a shape is produced  using the methods described in the following. Then, the fused image is created by multiplying the texture and the shape.

\begin{figure}
	\begin{center}
		  \includegraphics[width=0.8\linewidth]{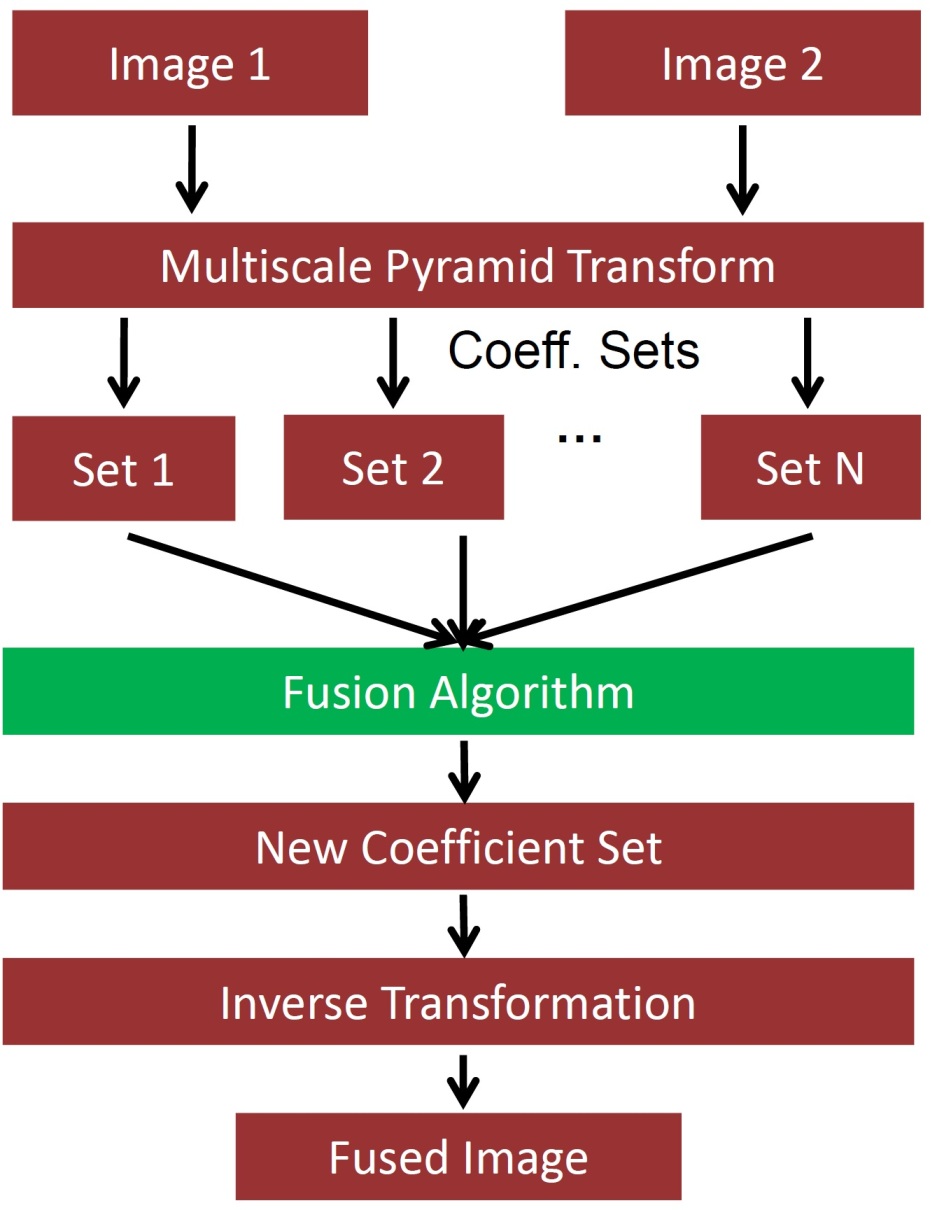}
		\caption{The pyramids algorithm for producing texture (the Fusion Algorithm is described in Figure~\ref{laplace})}
		\label{algorithm}
	\end{center}
\end{figure}

\subsection{Texture Production}

Image fusion using pyramids transformation is a useful method to fill in the gaps of one image using
another one(s). This method that was first proposed by Adelson \etal (1984) uses an iterative fusion scheme to
integrate information from multiple images~\cite{Adelson84}. Figure~\ref{algorithm} shows the algorithm
for this method. First, a filter is applied to the two input images to decompose them into sub-images 
(or equivalently to coefficients). Then, a fusion algorithm is applied to merge the subimages and produce a fused subimage. 
Finally, an inverse transformation will produce the merged image.

Rainfall is an intermittent event in space, so choose of the filter is one of the important aspects in this work. 
It is necessary to have a decomposition that can obtain the most information possible from the images.
Preliminary experiments using Gaussian, Laplacian, and Steerable pyramids showed that Steerable pyramids works better in
decomposing the images, and the resulting fused images are more accurate. Thus, I will only use the Steerable pyramids 
in the following. 

\begin{figure}
	\begin{center}
		  \includegraphics[width=0.8\linewidth]{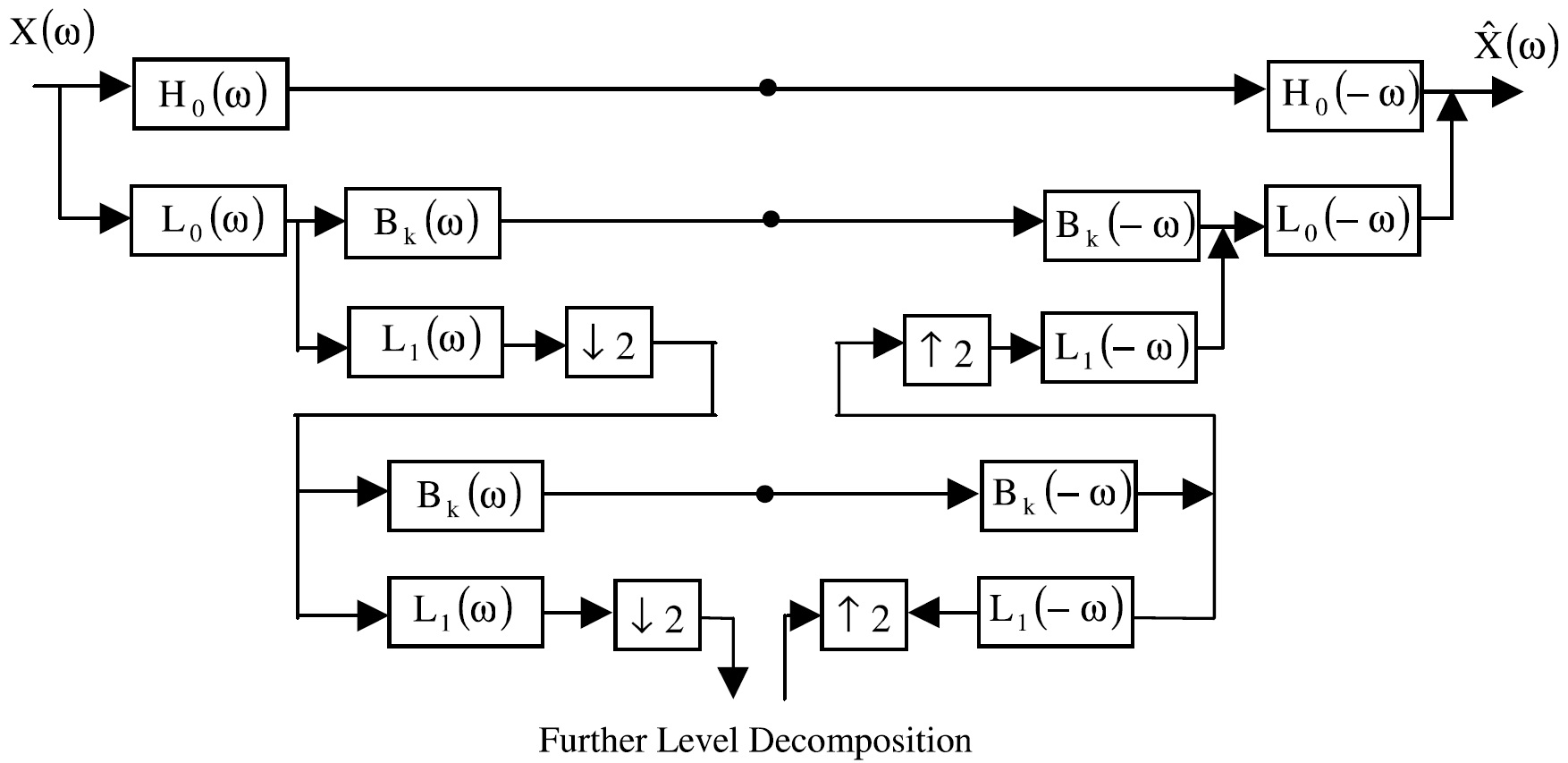}
		\caption{Structure of the Steerable pyramids in frequency domain (soruce: Liu \etal~\cite{Liu01})}
		\label{steer}
	\end{center}
\end{figure}

Steerable pyramids is a linear multi-scale, multi-orientation and self-inverting image decomposition introduced by Freeman and Adelson~\cite{Freeman91}.
Using Steerable filters $\mathit{L}_0$ and $\mathit{H}_0$, the image is first decomposed into low-pass and high-pass subbands. The low-pass band continues to be divided into a set of oriented bandpass subbands $\mathit{B}_0, \cdots , \mathit{B}_N$ and a lower low-pass subband $\mathit{L}_1$. The lower low-pass subband is sub-sampled by a factor of 2 along the x and y directions. Repeating this procedure, further decomposition is achieved. Figure~\ref{steer} shows the structure of this pyramid. The steerable pyramid representation is translation-invariant and rotation-invariant.

Steerable pyramids has been widely used in vision problems and image analysis. Here, I use it to transform the initial images to a sub-image domain and inverse-transform the fused sub-images to the merged final image. Testing different levels and orientation of decomposition, I found that a decomposition level of 4 (which is the maximum possible) and a decomposition orientation of 16 is good enough to decompose the images.

\begin{figure}
	\begin{center}
		  \includegraphics[width=0.8\linewidth]{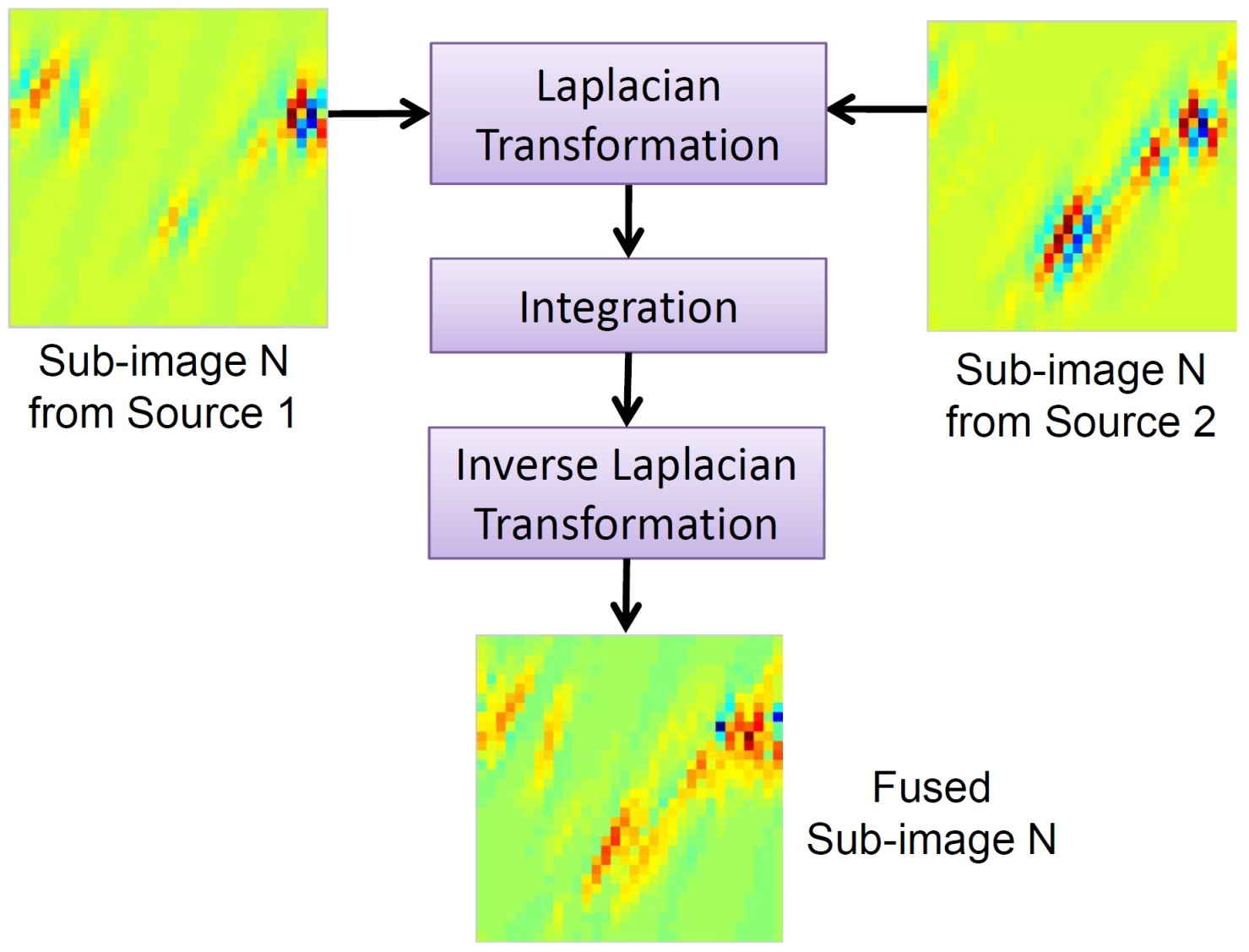}
		\caption{Fusion algorithm in the decomposed domain}
		\label{laplace}
	\end{center}
\end{figure}

As it was shown in Figure~\ref{algorithm}, a fusion algorithm is needed to combine the two sub-images (or coefficient sets) in the transformed domain.
A simple criteria for this integration is absolute value maximum selection (AVMS). This criteria selects the larger absolute 
values in the sub-images as the corresponding value for the fused sub-image. Although this criteria works well in many applications, 
it is so simple and it might miss some of the details in the integration process. Liu \etal (2001) proposed a fusion 
algorithm using the Laplacian pyramids~\cite{Liu01}. In this algorithm each pair of subimages from the two input images are 
decomposed by using the Laplacian filters. Then, an integration scheme is applied to merge the two pyramids of the subimages.
This integration scheme can be AVMS or any other defined criteria. Next, the fused subimage is produced by inverse transformation 
of the integrated pyramids of the input subimages. This fusion algorithm allows for more information to be fused into the merged subimages, 
and for the rainfall images that are so intermittent is really necessary. Therefore, I use this fusion algorithm for merging the subimages. Figure~\ref{laplace} shows the structure of the fusion algorithm.

\subsection{Shape Production}

This part of the algorithm will produce shapes for the rainfall fileds (\ie support). Early investigations on using the pyramids method to merge the two rainfall images revealed that the method will produce a kind of continuous texture in the resulting image, which is not the case in reality. Moreover, the pyramids method produces high values of false alarm in the fused images. Therefore, I implemented this part to constrain the rainy areas in specific regions of the final image. Nevertheless, these regions are still defined by the two input images, but using an interpolation scheme.

The procedure is that a linear interpolation is applied to create a merged image of the two input ones. Then, the pixels that were missing in both of the input images are set to be missing in the final image too. This is because of the high intermittency of rainfall patterns that makes it hard to interpolate the value of the missing pixels from the non-missing ones.

This part produces an image of values of [-1, 0, 1] in which -1 means missing pixel, 0 means no rain can occur, and 1 means rain can occur. Conditioning the texture to be only inside the possible rain areas, the fused measurement is produced.
\begin{figure}
	\begin{center}
		  \includegraphics[width=0.8\linewidth]{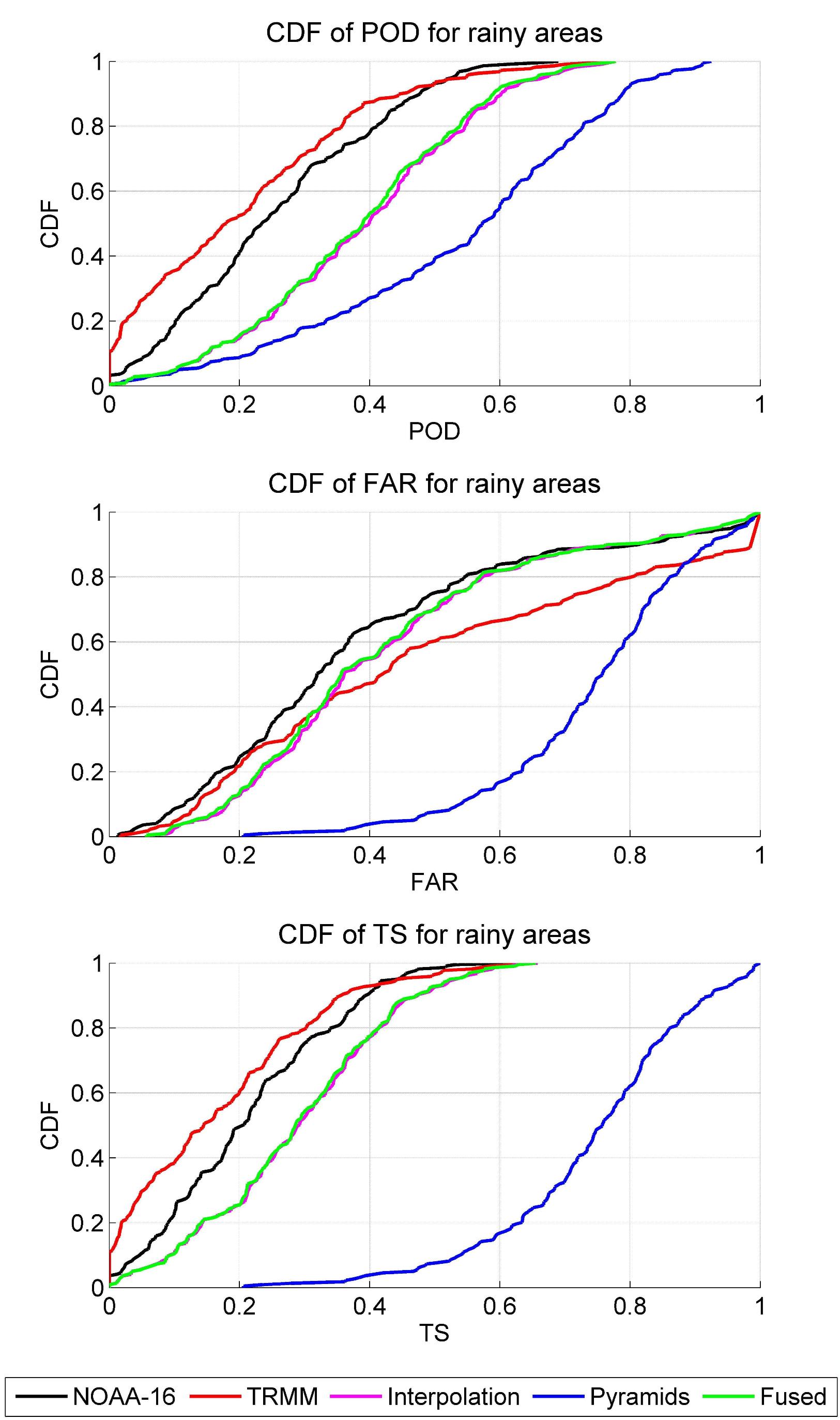}
		\caption{CDF plot of detection statistics for NOAA-16 and TRMM dataset}
		\label{N16}
	\end{center}
\end{figure}
\subsection{Fused Image Production}

This step is a multiplication of the produced texture and the produced shape. Therefore, the regions that have a shape value equal to 1 will have their values from the texture product, the regions that have a shape value equal to zero will be zero in the fused image, and the regions that their shape value is -1 (missing data) will remain missing in the fused image.

\section{Results}

In this section the results of the proposed algorithm are presented and evaluated for both rain detection and rain intensity. Ebert (2007) provides a comprehensive review on the statistics useful for validating the results of a rainfall estimation methods~\cite{Ebert07}. Here, I selected three statistics to evaluate the capability of the algorithm in detecting rainfall. The statistics are: Probability of Detection (POD), False Alarm Ratio (FAR), and Threat Score (TS). They are defined as following:

\begin{equation}
	POD = \frac{hits}{hits + misses}
\end{equation}

\begin{equation}
	FAR = \frac{false\; alarms}{hits + false\; alarms}
\end{equation}

\begin{equation}
	TS = \frac{hits}{hits + misses + false \;alarms}
\end{equation}

in which \emph{hits} is defined as the number of instances that both of the true measurement and the proposed measurement detect rain; \emph{misses} is the number of instances that the true detects rain, but the proposed one does not; and \emph{false alarms} is defined as the number of instances that the true does not detect rain, but the proposed one does. Note that I have only considered the detection of rain. Equivalently, it is possible to check the score for detection of no-rain, but as I consider both POD and FAR of rain it is not necessary to evaluate the no-rain detection score.

\begin{figure}
	\begin{center}
		  \includegraphics[width=0.8\linewidth]{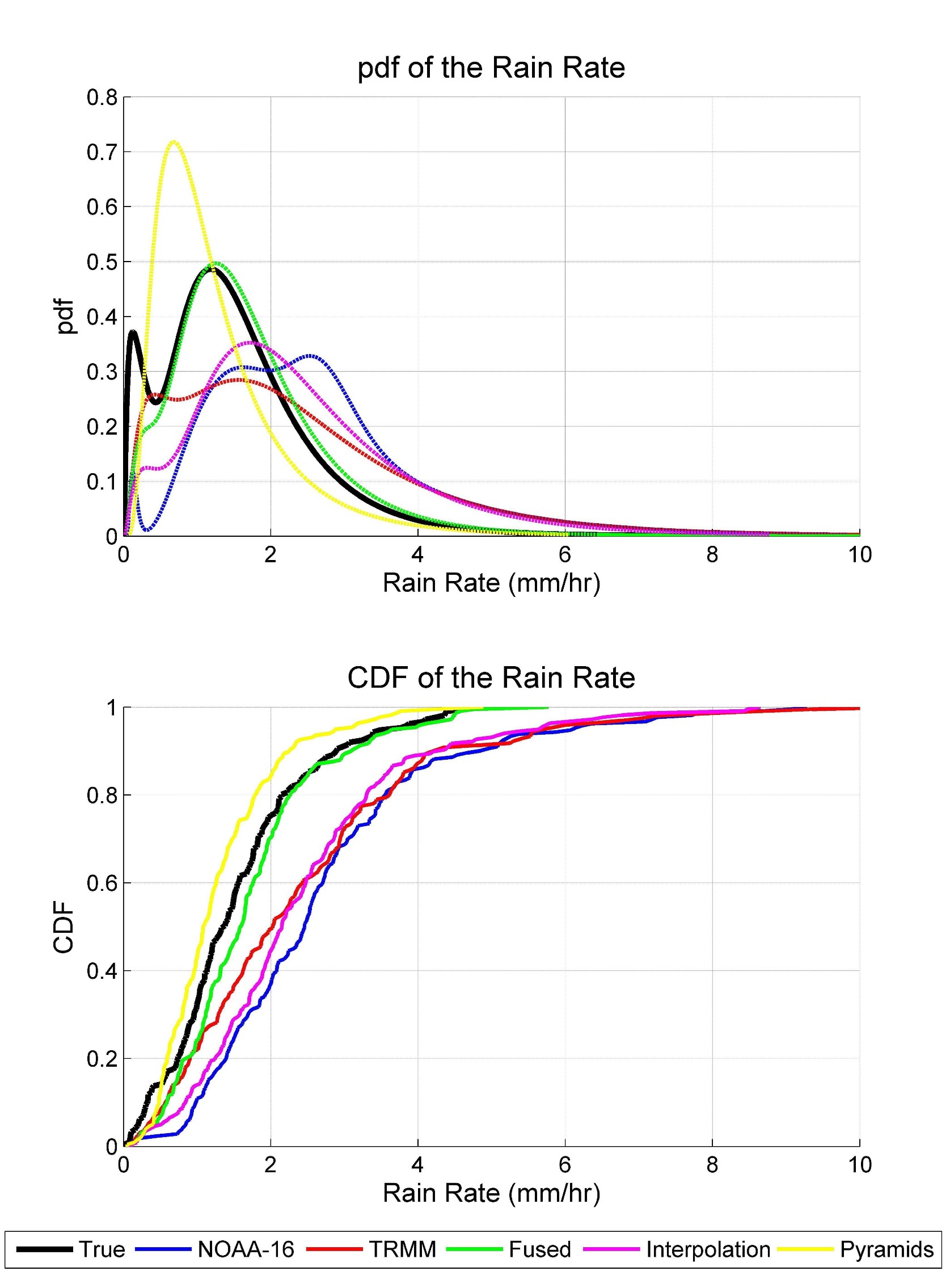}
		\caption{PDF and CDF of rain intensity for NOAA-16 and TRMM dataset}
		\label{cdfpdf}
	\end{center}
\end{figure}

\begin{figure}
	\begin{center}
		  \includegraphics[width=0.8\linewidth]{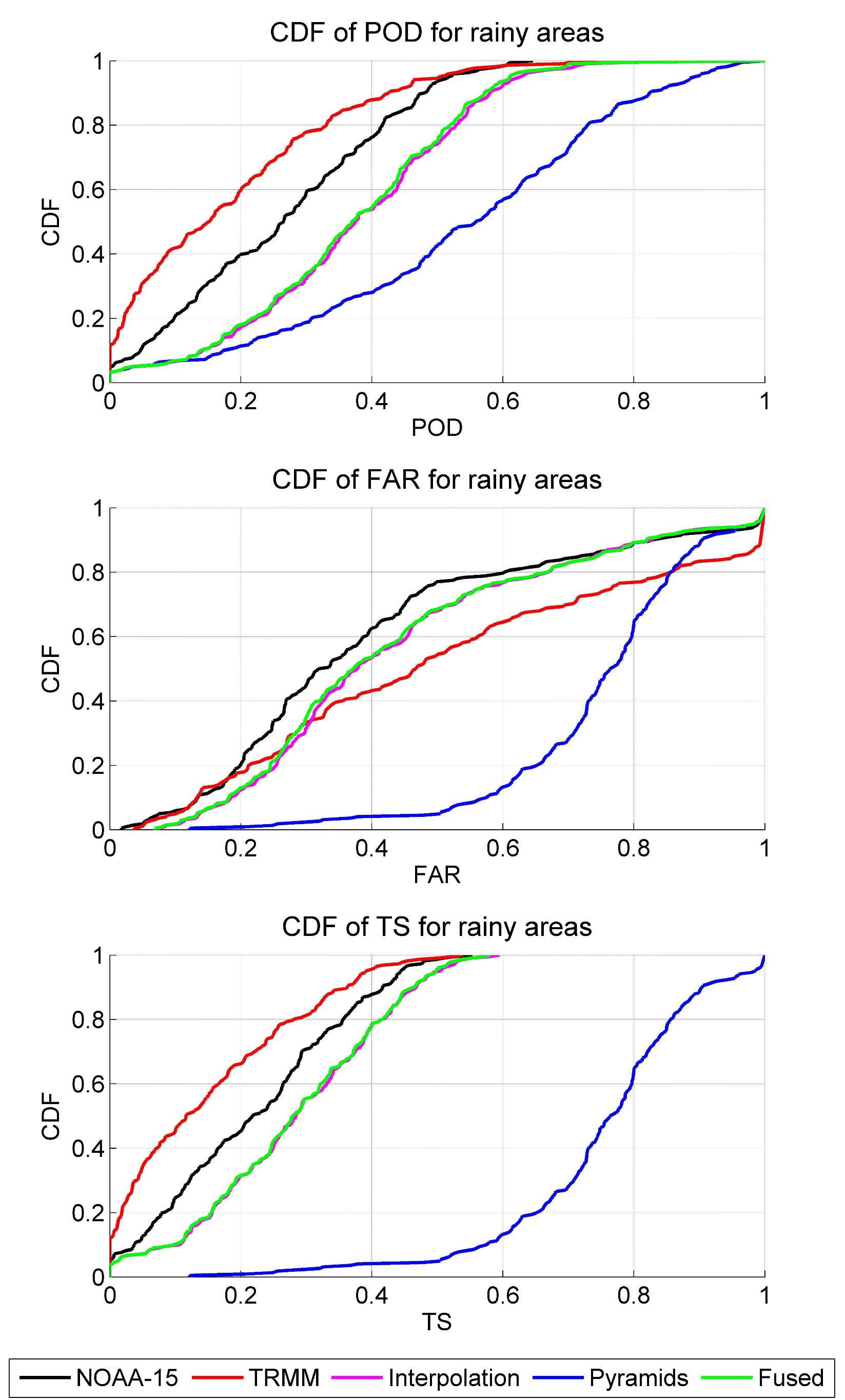}
		\caption{CDF plot of detection statistics for NOAA-15 and TRMM dataset}
		\label{N15}
	\end{center}
\end{figure}

The preferred values for each of these statistics are POD = 1, FAR = 0, and TS = 1. Figure~\ref{N16} shows the Cumulative Distribution Function (CDF) of the above mentioned statistics for NOAA-16 and TRMM dataset (with the characteristics described in section~\ref{dataset}) using the developed algorithm in this paper, the single interpolation method, and the pyramids method. As this figure shows, the pyramids method works well in improving the POD and TS, but at the same time it worsens the FAR to a high degree. Therefore, it is detecting the rainy areas by overestimating the rainy regions and giving too much false alarm that is not acceptable. The interpolation method seems to have a well matched result with the fused algorithm. They both improve the POD and TS with respect to the two input measurements (NOAA-16 and TRMM). Moreover, they provide a reasonable FAR which is somehow similar to the two input images. However, the difference between the interpolation method and the fused method becomes clear by comparing the Probability Distribution Function (PDF) and CDF of the rain intensity of different methods. This is an important comparison because a method that is only capable of detecting rainfall and does not provide reasonable intensities will not be useful. Figure~\ref{cdfpdf} shows the PDF and CDF of rain intensity for NOAA-16 and TRMM dataset. Note, these plots are for the pixels that were non-missing in all the measurements (\ie True, NOAA-16, TRMM, Fused, Interpolation, Pyramids.)

As can be seen in Figure~\ref{cdfpdf}, the PDF and CDF of the fused measurement match well with the ones for true measurement, which is the surface-based radar measurement as discussed in section~\ref{dataset}. In addition, to test which of these CDFs are from the same distribution as the true one, I run the Kolmogorov-Smirnov test~\cite{Massey51}. This test evaluates the hypothesis that two given datasets are from the same continuous distribution. The result with 5\% significance level showed that only the fused measurement can be from the same distribution as the true one. This shows that use of the fused algorithm, proposed in this paper, improves each of the pyramids and interpolation methods and provides better estimation with respect to both rain detection and rain intensity. 

\begin{figure}
	\begin{center}
		  \includegraphics[width=0.8\linewidth]{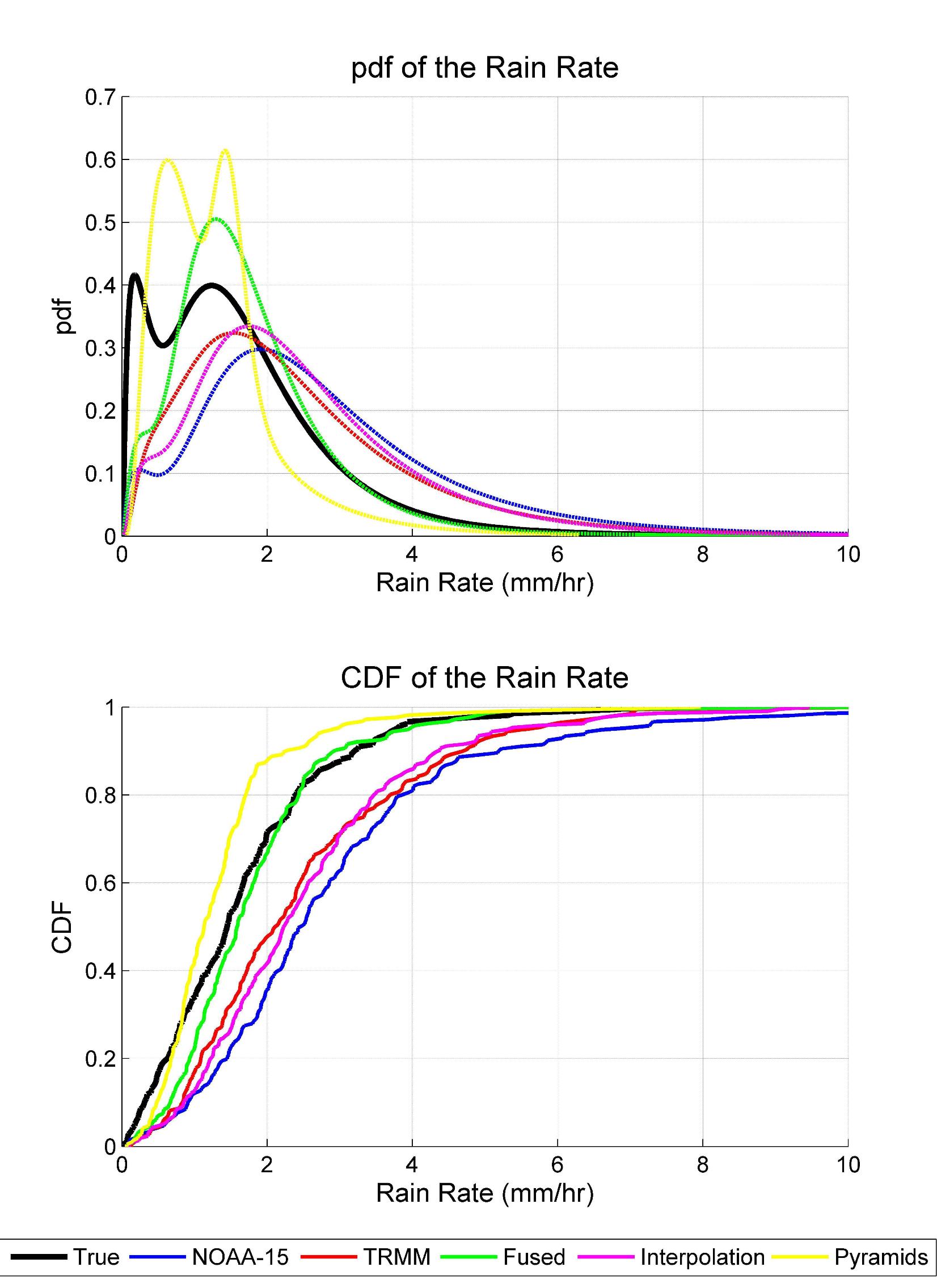}
		\caption{PDF and CDF of rain intensity for NOAA-15 and TRMM dataset}
		\label{cdfpdf15}
	\end{center}
\end{figure}

In order to further test the method, I run the method on another dataset which was the NOAA-15 and TRMM dataset (with the characteristics described in section~\ref{dataset}.) The results are illustrated in Figures~\ref{N15}~and~\ref{cdfpdf15}. As can be seen in these two images, the proposed fusion algorithm produces good results in this case too. The Kolmogorov-Smirnov test again showed that only the fused measurement CDF can be from the same distribution as the true one, at 5\% significance level. This proves the capability of the proposed algorithm in producing better estimations of rainfall for other satellites too.

\begin{figure}
	\begin{center}
		  \includegraphics[width=0.95\linewidth]{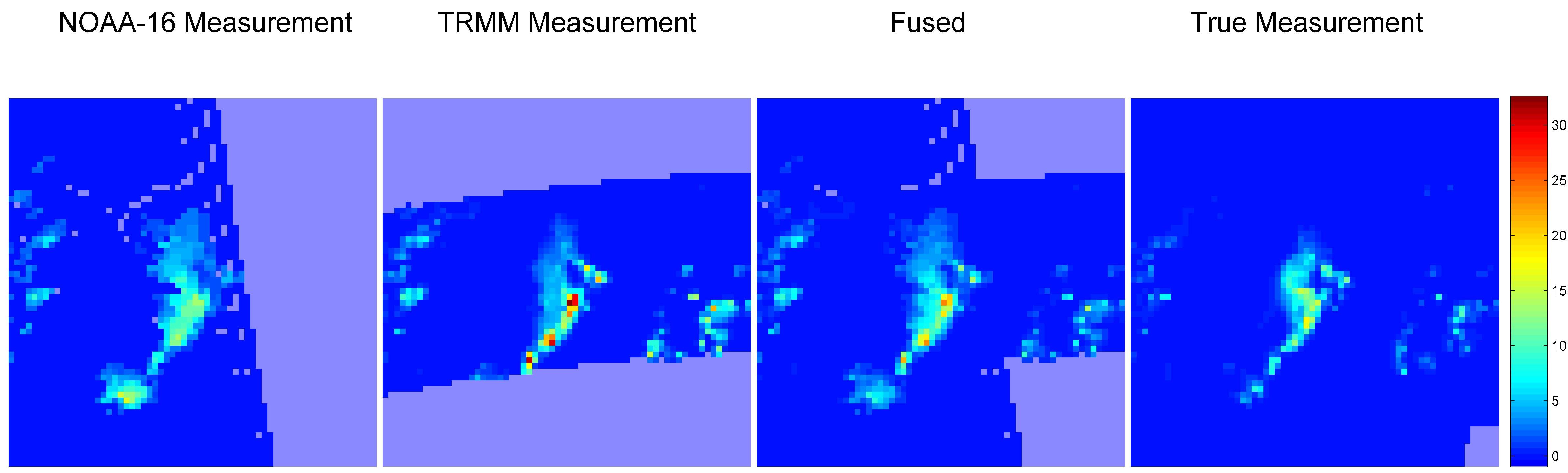}
		  \includegraphics[width=0.95\linewidth]{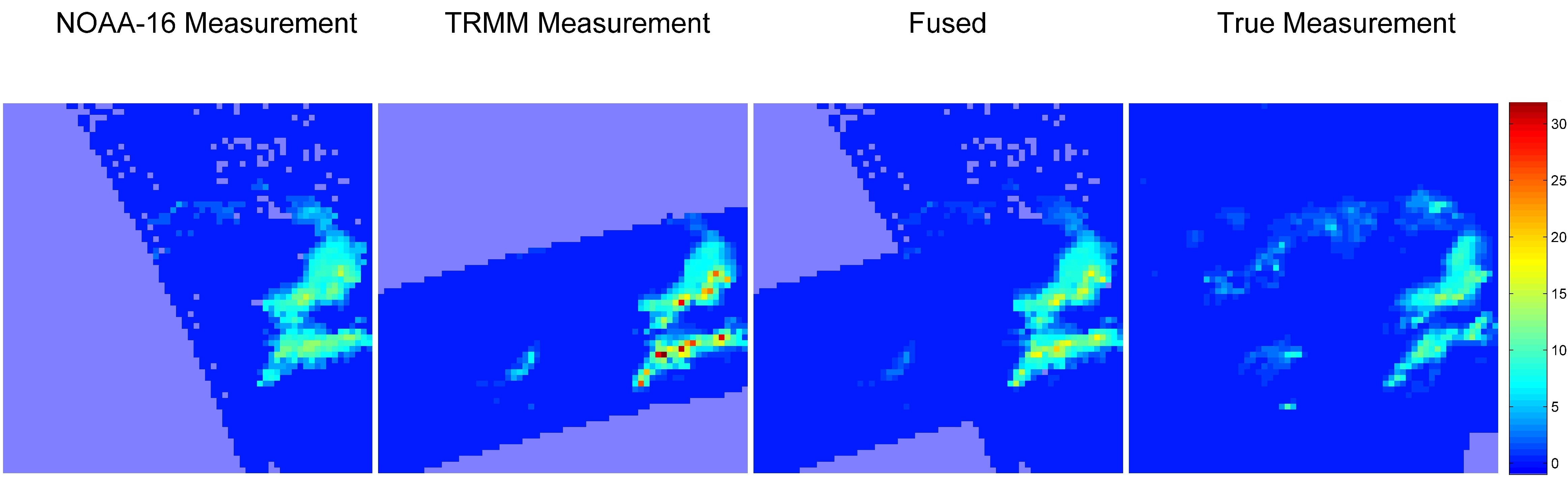}
		   \caption{Sample source, fused, and true measurements of NOAA-16 and TRMM dataset (the scale shows rain intensity in mm/hr)}
			\label{S1}
	\end{center}
\end{figure}

Figures~\ref{S1}~and~\ref{S2} show four samples of the sources, fused, and true images of rainfall intensity in NOAA-16 and NOAA-15 dataset. The purple color in these images represents the missing regions in the images. 

\section{Discussion}

This paper explored the possibility of using a multi-scale imagery technique in accordance with an interpolation scheme to merge two satellite measurements of rainfall and produce a better estimation of rain intensity with less missing points in the fused image. The proposed algorithm improves both rain intensity and rain detection of the two input images. Moreover, it was shown that only the results of the rain intensity from the fused image can be considered to be from the same distribution as the true measurement.

\begin{figure}
	\begin{center}
		  \includegraphics[width=0.95\linewidth]{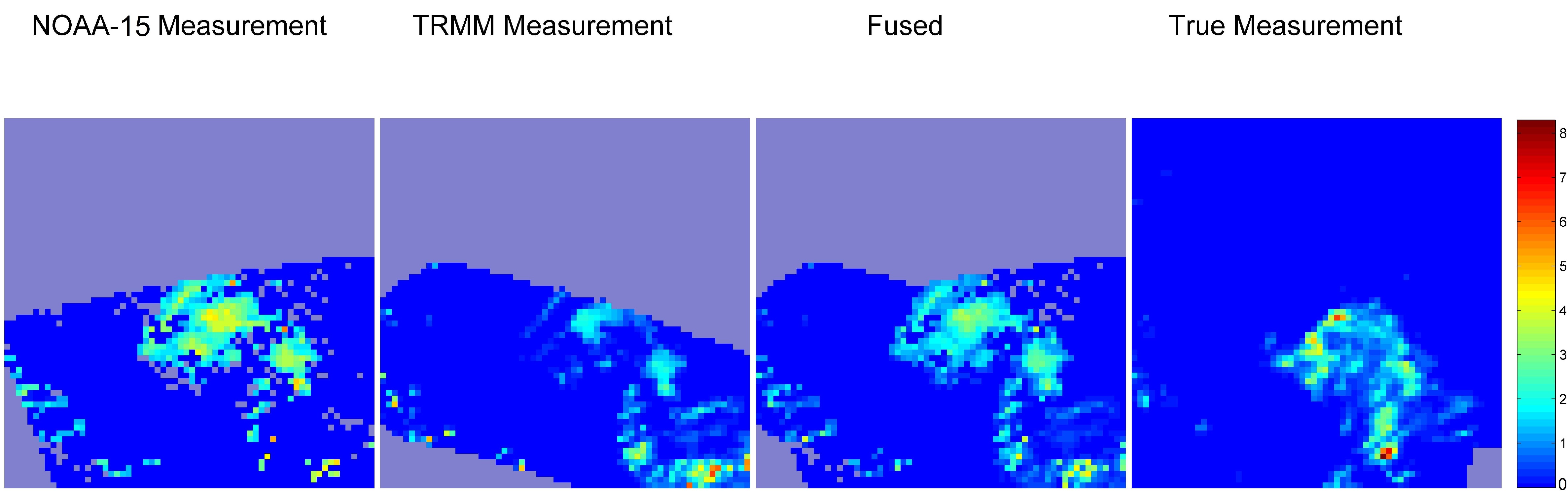}
		  \includegraphics[width=0.95\linewidth]{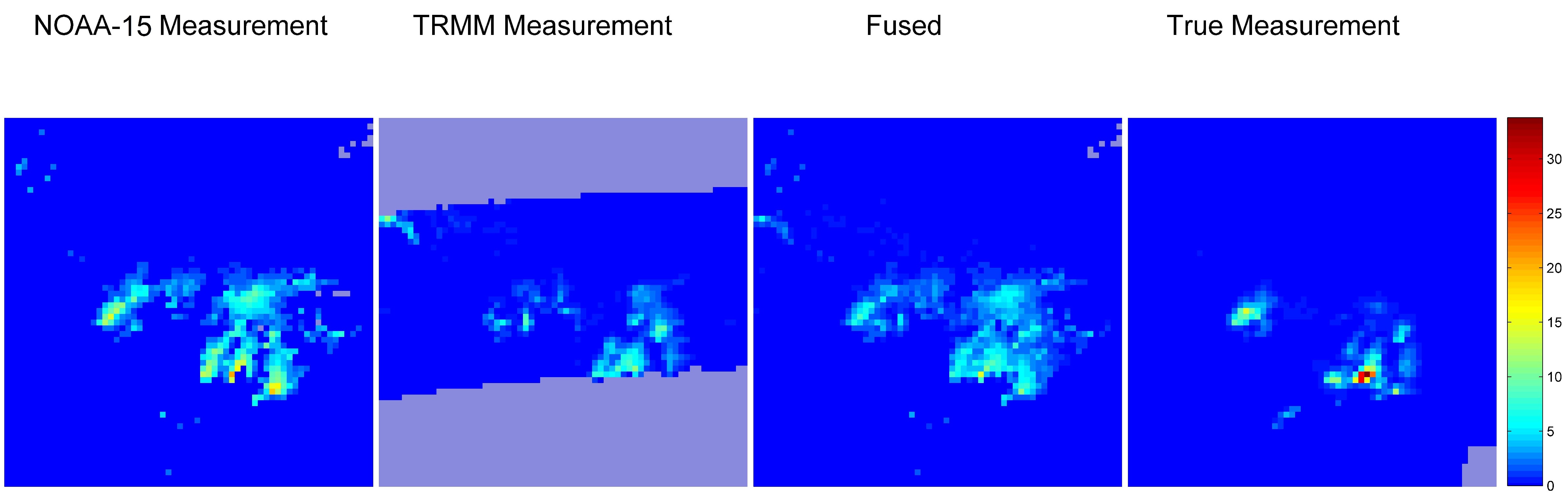}
		   \caption{Sample source, fused, and true measurements of NOAA-15 and TRMM dataset (the scale shows rain intensity in mm/hr)}
			\label{S2}
	\end{center}
\end{figure}

In addition, as can be seen in Figure~\ref{S2}, there are regions that there is no true measurement, and this algorithm can improve the satellite measurements and provide accurate estimates over them. Indeed, other the 50 sates of the US which have full surface-based radar coverage there are very few regions in the world that have a similar full coverage. Therefore, this method can be used in a any place over the globe to fuse different satellite measurements and produce better estimations of rainfall.

Furthermore, daily estimates of satellite rainfall can be improved by using this algorithm. There are places on the world that are not covered in the daily passes of one satellite, but they are covered by another satellites which miss some regions in the first one. The proposed algorithm can be applied to these daily measurements to provide full daily rainfall estimates that are useful for water balance studies and rainfall forecasting.

{\small
\bibliographystyle{ieee}
\bibliography{egbib}

\begin{thebibliography}{10}\itemsep=-1pt

\bibitem{Adelson84}
E.~H. Adelson, C.~H. Anderson, J.~R. Bergen, P.~J. Burt, and J.~M. Ogden.
\newblock Pyramid methods in image processing.
\newblock {\em RCA Engineer}, 29(6):33--41, 1984.

\bibitem{Ebert07}
E.~E. Ebert.
\newblock Methods for verifying satellite precipitation estimates.
\newblock In V.~Levizzani, P.~Bauer, and F.~J. Turk, editors, {\em Measuring
  precipitation from space}, pages 345--356. Springer, Dordrecht, 2007.

\bibitem{Freeman91}
W.~T. Freeman and E.~H. Adelson.
\newblock The design and use of steerable filters.
\newblock {\em IEEE Trans. Pattern Anal. Mach. Intell.}, 13(9):891--906, 1991.

\bibitem{Grimes08}
D.~I. Grimes.
\newblock An ensemble approach to uncertainty estimation for satellite-based
  rainfall estimates.
\newblock In S.~Sorooshian, K.~L. Hsu, E.~Coppola, B.~Tomassetti,
  M.~Verdecchia, and G.~Visconti, editors, {\em Hydrological Modeling and the
  Water Cycle}, pages 145--162. Springer, Berlin, 2008.

\bibitem{Hsu08}
K.~Hsu and S.~Sorooshian.
\newblock Satellite-based precipitation measurement using persiann system.
\newblock In S.~Sorooshian, K.~L. Hsu, E.~Coppola, B.~Tomassetti,
  M.~Verdecchia, and G.~Visconti, editors, {\em Hydrological Modeling and the
  Water Cycle}, pages 27--48. Springer, Berlin, 2008.

\bibitem{Joyce10}
R.~J. Joyce, P.~Xie, Y.~Yarosh, J.~E. Janowiak, and P.~A. Arkin.
\newblock Cmorph: A ''morphing" approach for high resolution precipitation
  product generation.
\newblock In M.~Gebremichael and F.~Hossain, editors, {\em Satellite Rainfall
  Applications for Surface Hydrology}, pages 23--37. Springer, Dordrecht, 2010.

\bibitem{Lensky08}
I.~M. Lensky and V.~Levizzani.
\newblock Estimation of precipitation from space-based platforms.
\newblock In S.~Michaelides, editor, {\em Precipitation: Advances in
  Measurement, Estimation and Prediction}, pages 195--217. Springer-Verlag,
  Berlin, 2008.

\bibitem{Liu01}
Z.~Liu, K.~Tsukada, K.~Hanasaki, Y.~K. Ho, and Y.~P. Dai.
\newblock Image fusion by using steerable pyramid.
\newblock {\em Pattern Recognition Letters}, 22:929--939, 2001.

\bibitem{Massey51}
F.~J. Massey.
\newblock The kolmogorov-smirnov test for goodness of fit.
\newblock {\em Journal of the American Statistical Association},
  46(253):68--788, 1951.

\bibitem{Shen10}
Y.~Shen, A.~Xiong, Y.~Wang, and P.~Xie.
\newblock Performance of high-resolution satellite precipitation products over
  china.
\newblock {\em J. Geophys. Res.}, 115:D02114, 2010.

\bibitem{Xie07}
P.~Xie, P.~A. Arkin, and J.~E. Janowiak.
\newblock Cmap: The cpc merged analysis of precipitation.
\newblock In V.~Levizzani, P.~Bauer, and F.~J. Turk, editors, {\em Measuring
  precipitation from space}, pages 319--328. Springer, Berlin, 2007.

\end{thebibliography}
}

\end{document}